\documentclass[10pt,conference]{IEEEtran}
\IEEEoverridecommandlockouts
\usepackage{cite}
\usepackage{amsmath,amssymb,amsfonts}
\usepackage{algorithmic}
\usepackage{graphicx}
\usepackage{textcomp}
\usepackage{xcolor}
\usepackage{orcidlink}
\def\BibTeX{{\rm B\kern-.05em{\sc i\kern-.025em b}\kern-.08em
T\kern-.1667em\lower.7ex\hbox{E}\kern-.125emX}}

\usepackage{blindtext}
\usepackage{verbatim}
\usepackage{url}
\usepackage{svg}
\usepackage{multirow}
\usepackage{subcaption}
\usepackage{graphicx}
\usepackage{tikz}
\usepackage{nicefrac}
\usetikzlibrary{quantikz}
\usetikzlibrary{shapes.geometric}
\tikzset{slice/.append style={line width=1.5pt}}

\usepackage{standalone}
\usepackage{tikzit}

\usepackage{xparse}

\newcounter{chatlinenum}

\definecolor{mygreen}{HTML}{88EABB}

\def\chatline#1\par{%
   \stepcounter{chatlinenum}%
   \noindent
   \ifodd\thechatlinenum
       \tikz[]{\node[fill=lightgray,chatstyle]{\strut#1\strut};}%
   \else
       \hfill
       \tikz[]{\node[fill=mygreen,chatstyle,align=right]{\strut#1\unskip\strut};}%
   \fi
   \par
   \smallskip
}

\begingroup
    \lccode`~=`\^^M
    \lowercase{%
\endgroup
    \def\newchatline#1~{%
        \stepcounter{chatlinenum}%
        \ifodd\thechatlinenum
            \tikz[]{\node[fill=lightgray,chatstyle]{\strut#1\strut};}%
        \else
            \hfill
            \tikz[]{\node[fill=mygreen,chatstyle,align=right]{\strut#1\strut};}%
        \fi
        ~
        \smallskip
    }%
}

\NewDocumentEnvironment{newchat}{}{%
    \setcounter{chatlinenum}{0}
    \begin{minipage}{\columnwidth}
        \obeylines
        \everypar={\newchatline}
}{%
    \end{minipage}
}

\tikzset{chatstyle/.style={text width=0.9\columnwidth, rounded corners=3pt}}

\usepackage{wasysym}
\usepackage{textcomp}
\usepackage{makecell}

\begin{document}

\title{Applying QNLP to sentiment analysis in finance}


\makeatletter
\newcommand{\linebreakand}{%
  \end{@IEEEauthorhalign}
  \hfill\mbox{}\par
  \mbox{}\hfill\begin{@IEEEauthorhalign}
}
\makeatother

\author{
\IEEEauthorblockN{Jonas Stein\textsuperscript{$\orcidlink{0000-0001-5727-9151}$}}
\IEEEauthorblockA{\textit{LMU Munich}}
\and
\IEEEauthorblockN{Ivo Christ}
\IEEEauthorblockA{\textit{Aqarios GmbH}}
\and
\IEEEauthorblockN{Nicolas Kraus}
\IEEEauthorblockA{\textit{Aqarios GmbH}}
\linebreakand 
\IEEEauthorblockN{Maximilian Blathasar Mansky}
\IEEEauthorblockA{\textit{LMU Munich}}
\and
\IEEEauthorblockN{Robert Müller\textsuperscript{$\orcidlink{0000-0003-3108-713X}$}}
\IEEEauthorblockA{\textit{LMU Munich}}
\and
\IEEEauthorblockN{Claudia Linnhoff-Popien\textsuperscript{$\orcidlink{0000-0001-6284-9286}$}}
\IEEEauthorblockA{\textit{LMU Munich}}
}

\maketitle

\begin{abstract}
As an application domain where the slightest qualitative improvements can yield immense value, finance is a promising candidate for early quantum advantage. Focusing on the rapidly advancing field of Quantum Natural Language Processing (QNLP), we explore the practical applicability of the two central approaches DisCoCat and Quantum-Enhanced Long Short-Term Memory (QLSTM) to the problem of sentiment analysis in finance. Utilizing a novel ChatGPT-based data generation approach, we conduct a case study with more than 1000 realistic sentences and find that QLSTMs can be trained substantially faster than DisCoCat while also achieving close to classical results for their available software implementations.

\end{abstract}


\section{Introduction}
\label{sec:introduction}
Predicting the price development of assets is one of the core problems in finance. As the price of any asset is determined by supply and demand, the individual valuation of each buyer and seller constitutes essential information for the task of price prediction. With the advent of social media platforms like twitter, such data became freely available at massive scale. This catalyzed the application of methods from natural language processing (NLP) to allow for automated analysis of investor's sentiments~\cite{8455771}. While the field of NLP has achieved tremendous success in recent history~\cite{info14040242} -- even showing substantial evidence to pass the Turing test~\cite{10.1093/mind/LIX.236.433, Nath2022} -- the compute used to achieve these advancements is enormous~\cite{10.1145/3458817.3476209}.

A particularly promising novel approach to NLP and computational finance in terms of hardware requirements is quantum computing, as it has been shown to need less training data in machine learning \cite{caro2022generalization} and allows for the efficient incorporation of grammatical structure in NLP ~\cite{app12115651, Stamatopoulos2022towardsquantum}. Motivated by recent positive experimental results of two quantum NLP (QNLP) methods, namely the Quantum-Enhanced Long Short-Term Memory (QLSTM) neural networks~\cite{9747369, DiSipio.2022} and the quantum native DisCoCat (Distributional Compositional Categorical) based QNLP~\cite{Meichanetzidis2020, 10.1613/jair.1.14329, Martinez.2022}, we evaluate their applicability to sentiment analysis on financial datasets beyond existing proofs of concept.


This work encompasses the following contributions:
\begin{itemize}
    \item A comprehensive guide for the application of QLSTMs and DisCoCat.
    \item Introducing a novel approach to generating datasets for DisCoCat that enables automated data generation with sufficient grammatical structure using ChatGPT.
    \item A case study comparing the applicability of two state-of-the-art QNLP methods in a finance application.
\end{itemize}

The remainder of this paper is structured as follows. Sec.~\ref{sec:background} discusses theoretic preliminaries on QLSTMs and DisCoCat. Sec.~\ref{sec:methodology} shows how both methods can be applied to solve the given task of sentiment analysis. Sec.~\ref{sec:evaluation} contains the evaluation and subsequent comparison of both approaches on synthetic datasets. Sec.~\ref{sec:conclusion} concludes the findings.

\section{Background}
\label{sec:background}
A central component in state-of-the-art NLP approaches is choosing how to model text mathematically using \emph{distributional semantics}, i.e., mapping each word onto a vector, which generally leads to similar words being described by vectors with a high similarity~\cite{KHATTAK2019100057}. Historically, such \emph{word embeddings} where mainly done in a static manner, meaning that words were mapped onto vectors based on their possible meanings according to, e.g., dictionaries~\cite{10.5555/555733}. As NLP evolved, more context aware word embeddings were used, which allows solving potential problems raised from \emph{homonyms}~\cite{kimura1992association}. Homonyms are words that have different meanings based on the context they are used in, e.g., the word ``rock'' can reference a stone or a genre of music. Today's state-of-the-art approaches in regards to qualitative performance avoid such problems using \emph{dynamic} embeddings, which is heavily driven by the availability of massive amounts of data and immense computing resources~\cite{openai2023gpt4}.

Expanding on the mathematical model for text, we now examine methods to process it computationally. As text is a form of temporal data with variable length, recurrent neural networks (RNNs) have been extensively used to infer information from the text~\cite{tarwani2017survey}. However, ordinary RNNs have difficulties accounting for a  grammatical semantics that extend beyond locally available information in a sentence~\cite{tang2018self}, e.g., as in ``The view from this ugly skyscraper was incredibly beautiful'', where the semantically related words ``view'' and ``beautiful'' are separated by almost the whole sentence. Long term dependencies are successfully addressed in a special type of RNNs, i.e., LSTMs, by using the propagation of a cell state in addition to the hidden state over time~\cite{10.1162/neco.1997.9.8.1735}. Expanding on this, the concept of transformers was introduced, which focus on self-attention while shifting away from the recurrent network structure to allow for parallelization in the otherwise necessarily iterative training procedure~\cite{Vashwani2017}.

In combination with substantial compute and large amounts of data, current high performance NLP approaches have shown the ability to learn grammatical semantics without any explicit knowledge about grammar. A completely different approach from RNNs, LSTMs and Transformers is DisCoCat, which natively integrates grammatical information given in \emph{syntax trees}\footnote{In the context of grammar, a syntax tree describes the parts of speech such as nouns, verbs, adjectives, etc., and puts them in structural context~\cite{lambek2008word}.} using \emph{category theory} to unify distributional semantics with the \emph{principle of compositionality}\footnote{This principle states that the meaning of an expression is fully described by the meaning and composition of its parts~\cite{coecke2010mathematical}.}. In essence, DisCoCat exploits a mathematical similarity between grammar, expressed through syntax trees, and quantum computing~\cite{Meichanetzidis2020}.


Expanding on this brief overview of NLP, we now show how quantum computing techniques can be productively applied to enhance LSTMs and DisCoCat based NLP. 

\subsection{Using QLSTMs for QNLP}\label{subsec:QLSTMs}
(Q)LSTMs can be viewed as an extension of conventional (Q)RNNs with two core differences: (1) the introduction of an additional cell state $c_t$ for long term memory storage besides the hidden state $h_t$ and (2), the replacement of the single neural network with a specific composition of neural networks (i.e., a (Q)LSTM \emph{cell}) as shown in Fig.~\ref{fig:qlstm}.~\cite{10.1162/neco.1997.9.8.1735, 9747369}
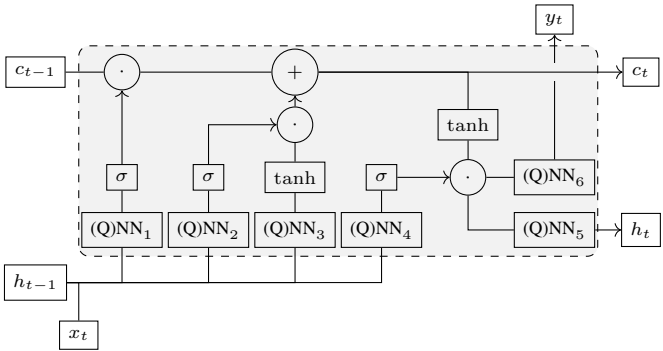
\begin {figure}[htbp]
\centering
\resizebox{\columnwidth}{!}{\begin{tikzpicture}[every node/.style={font=\scriptsize},
gate/.style={rectangle, draw=black}, connection/.style={circle, draw=black},
xscale=1.15, yscale=.7]
\draw (0,0) node [gate] (cell-state) {$c_{t-1}$} (0, -4) node [gate] (hidden-state) {$h_{t-1}$};
\draw (1,0) node [connection] (cell-mult) {$\cdot$} (1,-2) node [gate] (cell-activation) {$\sigma$} (1, -3) node [gate] (qnn1) {$\text{(Q)NN}_1$};
\draw (2,-2) node [gate] (cell-activation-2) {$\sigma$} (2, -3) node[gate] (qnn2) {$\text{(Q)NN}_2$};
\draw (3, 0) node [connection] (cell-plus) {$+$} (3,-1) node[connection] (cell-mult-2) {$\cdot$} (3,-2) node[gate] (cell-tanh) {$\tanh$} (3,-3) node[gate] (qnn3) {$\text{(Q)NN}_3$};
\draw (4,-2) node [gate] (cell-activation-3) {$\sigma$} (4,-3) node[gate] (qnn4) {$\text{(Q)NN}_4$};
\draw (5,-1) node [gate] (cell-tanh-2) {$\tanh$} (5,-2) node[connection] (hidden-mult) {$\cdot$};
\draw (6,1) node[gate] (output) {$y_t$} (6,-2) node [gate] (qnn6) {$\text{(Q)NN}_6$} (6, -3) node[gate] (qnn5) {$\text{(Q)NN}_5$};
\draw (7,0) node[gate] (cell-output) {$c_t$} (7,-3) node[gate] (hidden-output) {$h_t$};
\draw (0.5, -5) node[gate] (input) {$x_t$};

\draw[->] (qnn1) -- (cell-activation) -- (cell-mult);
\draw[->] (qnn2) -- (cell-activation-2) |- (cell-mult-2);
\draw[->] (qnn3) -- (cell-tanh) -- (cell-mult-2) -- (cell-plus);
\draw[->] (qnn4) -- (cell-activation-3) -- (hidden-mult);
\draw[->] (hidden-mult) |- (qnn5) -- (hidden-output);
\draw (hidden-mult) -- (cell-tanh-2) |- (cell-plus);
\draw[->] (hidden-mult) -- (qnn6) -- (output);
\draw (6,0) node[rectangle, fill=gray!10] (overlap) {};
\draw[->](cell-state) -- (cell-mult) -- (cell-plus) -- (cell-output);
\draw (hidden-state) -| (qnn1)
  (hidden-state) -| (qnn2)
  (hidden-state) -| (qnn3)
  (hidden-state) -| (qnn4);
\draw (input) |- (hidden-state);
\begin{scope}[on background layer]
\path[draw=black, dashed, rounded corners, fill=gray!10] (.5, -3.5) rectangle (6.5, .5);
\end{scope}
\end{tikzpicture}}
\caption{Structure of a (Q)LSTM in which the internal processing pipeline is indicated by arrows. Inputs are the cell state $c_{t-1}$ of the previous time step, the corresponding hidden state $h_{t-1}$ and the current state of the time series data $x_t$ to be processed. $c_t$ and $h_t$ denote the computed cell- and hidden states passed to the next cell. The output of the cell is $y_t$.}
\label{fig:qlstm}
\end{figure}

In this figure, $\sigma$ denotes a sigmoid activation function, analogue notation is used for $\tanh$. The operations $\bigodot$ and $\bigoplus$ denote element-wise multiplication and addition of vectors. $x_t$ denotes the data input at point $t$ and $y_t$ the output at the same point in time, $h_t$ denotes the hidden state similar to the hidden state in conventional RNNs and $c_t$ denotes the cell state of the LSTM. \cite{Goodfellow2016}

The LSTM cell is divided into three neural networks acting as gates:
\begin{enumerate}
    \item The \emph{forget gate}: The first (Q)NN determines the amount of information to be forgotten about the state $c_{t-1}$. The lower the value of $\sigma\left(\text{(Q)NN}_1\left(h_{t-1}, x_t\right)\right)$, the more the resulting entries of $c_{t-1}$ approach zero.
    \item The \emph{input gate}: Involving the second and third QNN, the update gate is split into deciding how much information is to be updated ((Q)NN${}_2$) and which information is to be updated  ((Q)NN${}_3$).
    \item The \emph{output gate}: Determines the information to be output based on all states $x_t$, $h_{t-1}$ and the processed $c_{t-1}$.
\end{enumerate}
In the presented structure, additional postprocessing of the generated state from the output gate is conducted individually to allow for yielding different outputs $h_t$ and $y_t$ (in some implementations, $y_t=h_t$ is used). Choosing suitable dimensionalities for $h$ and $c$, as well as architectures for all (Q)NNs, (Q)LSMTs can be used to process time series completely analog to standard (Q)RNNs~\cite{dhake2023algorithms}.

\subsection{Using DisCoCat for QNLP}\label{subsec:DisCoCat}
Following the original DisCoCat approach proposed by Coecke et al.~\cite{coecke2010mathematical, EPTCS221.8}, there are three steps to generate a mathematical model of a sentence that can be processed using quantum computers:
\begin{enumerate}
    \item \label{itm:step1} Parse the sentence into a pregroup expression.
    \item \label{itm:step2} Construct the corresponding DisCoCat diagram (incl. possible reductions).
    \item \label{itm:step3} Translate the DisCoCat diagram into a quantum circuit according to possibly parameterized word embeddings.
\end{enumerate}
A pregroup $\left(A, 1, \cdot, -^l, -^r,\leq\right)$ can be understood as a special non-commutative group, i.e., essentially a group with differentiating between left and right inverses and an order relation. By assigning a type (i.e., a pregroup element) to each word in a sentence and concatenating these with suitable left and right inverses depending on their grammatical function in the sentence, we can express the sentence in form of a pregroup expression. Considering the sentence ``Alice loves Bob'' and following the established pregroup grammar for the English language~\cite{EPTCS221.8}, all nouns are identified with the pregroup element $n$ and the verb ``loves'' is identified as $n^r \cdot s \cdot n^l$, meaning that it expects a noun from the left and right and yields a complete sentence $s$ when concatenated with these:
\begin{align}
    \text{``Alice loves Bob''} \quad \mapsto\quad  & n \cdot \left( n^r \cdot s \cdot n^l \right) \cdot n \label{eq:map-to-pregroup}\\
    =  &\left( n \cdot  n^r \right) \cdot s \cdot \left( n^l  \cdot n \right) = 1 \cdot s \cdot 1 = s \nonumber
\end{align}
Expanding on this model of grammar, we now introduce the distributional semantics aspect of DisCoCat to construct the DisCoCat diagram (step~\ref{itm:step2}). This is done by mapping each atomic type $a\in A$ onto a vector space, i.e., $n$ to $N$ and $s$ to $S$, and their concatenation to tensor product spaces ($ n^r \cdot s \cdot n^l$ to $N\otimes S\otimes N$). Therefore, a pregroup expression is then represented as a concatenation of functions, e.g., ``loves'' becomes a bilinear map $N\times N \rightarrow S$ while adjectives (of type $n\cdot n^l$) are represented by a linear map $N\rightarrow N$. Using the diagrammatic calculus of compact closed categories, such computations can be represented in so-called DisCoCat diagrams, as shown in Fig.~\ref{fig:discocat}. Following the established choices of maps and using complex vector spaces, these diagrams can be reshaped according to mathematical properties as, e.g., the isomorphism between row and column vectors through taking the conjugate transpose. A possible reduction of the example in Fig.~\ref{fig:discocat} is depicted in Fig.~\ref{fig:discocat-reduced}.

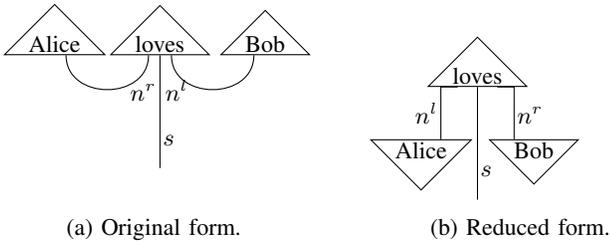
\begin{figure}[t]
  \begin{subfigure}{0.45\columnwidth}
    \begin{tikzpicture}[xscale=1.4, yscale=2,
every node/.style={font=\small, inner sep=1pt},
state/.style={isosceles triangle, draw, shape border rotate=90, isosceles triangle apex angle=90}, 
measurement/.style = {isosceles triangle, draw, shape border rotate=-90, isosceles triangle apex angle=90}]
\draw (0,0) node[state] (alice) {Alice}
	(1,0) node[state] (loves) {loves}
	(2,0) node[state] (bob) {Bob};
\draw (alice.south east) to[bend right=90] (loves.south west) node [below, yshift=-8pt, xshift=-2pt] {$n^r$\vphantom{$n^l$}};
\draw (loves.south east) node[below, yshift=-8pt, xshift=2pt] {$n^l$} to [bend right=90] (bob.south west);
\draw (loves.south) --++ (0,-.75) node [near end, right] {$s$};
\end{tikzpicture}
    \caption{Original form.}
    \label{fig:discocat}
  \end{subfigure}
  \hfill
  \begin{subfigure}{0.45\columnwidth}
    \begin{tikzpicture}[xscale=.75, yscale=1,
every node/.style={font=\small, inner sep=1pt},
state/.style={isosceles triangle, draw, shape border rotate=90, isosceles triangle apex angle=90}, 
measurement/.style = {isosceles triangle, draw, shape border rotate=-90, isosceles triangle apex angle=90}]
  \draw (0,-1) node[measurement] (alice) {Alice}
	(1,0) node[state] (loves) {loves}
	(2,-1) node[measurement] (bob) {Bob};
\draw (alice.30) |- node[near start, left] {$n^l$} (loves.210) (loves.330) -| node[near end, right] {$n^r$\vphantom{$n^l$}} (bob.150);
\draw (loves.south) -- node [near end, right] {$s$}++ (0,-1.5) ;
\end{tikzpicture}
    \caption{Reduced form.}
    \label{fig:discocat-reduced}
  \end{subfigure} 
    \caption{DisCoCat diagrams for ``Alice loves Bob''.}
  \label{fig:bigone}
\end{figure}

To finally conduct step~\ref{itm:step3}, i.e., mapping the resulting DisCoCat diagram onto a quantum circuit, $\bigtriangledown$ boxes are substituted by quantum states $\ket{\psi}$ based on the chosen word embedding, while $\bigtriangleup$ boxes correspond to their adjoints $\bra{\psi}$. Reading the DisCoCat diagrams from top to bottom and defining straight wires $\mid$ as identity operations, this leads to the quantum circuit shown in Fig.~\ref{fig:circuit}. The different structures shown in equation~\eqref{eq:map-to-pregroup}, the diagrammatic representation in Fig.~\ref{fig:bigone} and the quantum circuit in Fig.~\ref{fig:circuit} are equivalent in a categorical sense.

Note that this QNLP approach requires classical postselection, i.e., to extract the meaning of the sentence $\ket{s}$, wires 1 and 3 have to be measured in the $\ket{0}^{\otimes \left| N\right|}$ state. In practice, this can contribute to a potentially immense increase in overall runtime, especially when scaling up the number of qubits per DisCoCat wire, i.e., in this case, the dimensionality of $N$.

\section{Methodology}
\label{sec:methodology}

In the following, we describe our process of data generation, the choice for the classical baseline facilitating meaningful evaluation of the results, as well as the implementation of the QLSTM and the DisCoCat approach.

\subsection{Dataset generation}
Aiming to use the same datasets for the QLSTM and DisCoCat approaches, the sentences must inherit a grammatical structure that is comprehensible to the syntax parser employed in DisCoCat. To the authors' awareness, no real world datasets inheriting this property while also being small enough for current quantum circuit simulators are available to date. Motivated by the recent success of large language models like ChatGPT~\cite{OpenAI2022}, we propose to employ such means to create suitable synthetic datasets. To generate simple sentences containing sentiments on finance topics, we interacted with ChatGPT in the following way:

\noindent\begin{newchat}
Generate sentences with a maximum length of five words discussing financial topics or stocks in a positive, neutral or negative way. At the end of each sentence, mention its respective label with negative being 0, neutral being 1 and positive being 2. \hfill $[$\textit{Query}$]$
Inflation fears rattle markets (Negative - 0) \\Interest rates stay steady (Neutral - 1)\\$[$\textit{ChatGPT}$]$ \hfill Apple reports record profits (Positive - 2)
\end{newchat}

To generate more complex statements, we altered the input to: ``Generate detailed sentences discussing financial topics or stocks in a positive, neutral or negative way. [...]''. An example of a positive reply to this was ``The rise of online banking has made it easier and more convenient for customers
to manage their finances''.

An overview of the generated datasets is displayed in Tab.~\ref{tab:DataDistribution}.

\begin{table}[h]
    \centering
    \begin{tabular}{c|c|c|c|c|c}
        & \multicolumn{3}{c|}{Class} & & \\ \hline
        Complexity & $-$ & $\circ$ & $+$ & $\diameter$ word count & vocabulary size \\ \Xhline{2\arrayrulewidth}
         Low & 34\% & 18\% & 48\% & 4.9 & 913 \\ \hline
         Moderate & 37\% & 17\% & 46\% & 18.4 & 1608
    \end{tabular}
    \caption{Data class distribution of both generated datasets encompassing roughly 1000 sentences each.}
    \label{tab:DataDistribution}
\end{table}

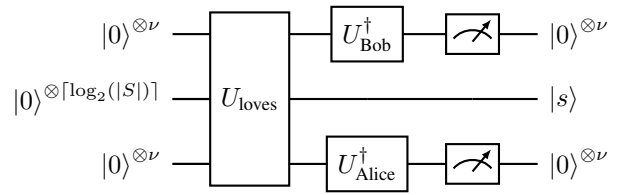
\begin {figure}[t]
\centering
\begin{quantikz}
\lstick{$\ket{0}^{\otimes\nu}$} & \gate[3]{U_{\text{loves}}} & \gate{U_{\text{Bob}}^{\dagger}} & \meter{} & \rstick{$\ket{0}^{\otimes \nu}$}\qw \\
\lstick{$\ket{0}^{\otimes \lceil\log_2\left(\left| S\right|\right)\rceil}$} & \qw & \qw & \qw & \rstick{$\ket{s}$}\qw   \\
\lstick{$\ket{0}^{\otimes\nu}$} & \qw & \gate{U_{\text{Alice}}^{\dagger}} & \meter{} & \rstick{$\ket{0}^{\otimes\nu}$}\qw 
\end{quantikz}
\caption{A quantum circuit for ``Alice loves Bob'' when using an arbitrary word embedding $U_{\text{word}}$, where $\nu \coloneqq  \lceil\log_2\left(\left| N\right|\right)\rceil$.} 
\label{fig:circuit}
\end{figure}

\subsection{Classical Baseline LSTM}
To allow for a meaningful comparison of results, we employ a LSTM as the classical baseline. Following Sec.~\ref{subsec:QLSTMs}, a hyperparameter search the low (moderate) complexity datasets lead to the use of the ReLU activation function, a trainable word embedding layer with size 5 (10), one unidirectional LSTM layer, one fully connected hidden layer of size 8 (16), followed by a dropout layer with rate 0 (0.1).

\subsection{QLSTM Implementation}
Analogously, QLSTM implementation also follows the general architecture described in Sec.~\ref{subsec:QLSTMs} but differs from the classic baseline by exchanging the LSTM with a QLSTM cell. As a result of a basic hyperparameter tuning for the low (moderate) complexity datasets, we choose: A three qubit one hot encoding for $y_t$, four qubits and one ansatz layer in each QNN, whereas the ansatz corresponds to that proposed in the original QLSTM paper\footnote{The ansatz consists of a data input layer, a variational layer and a measurement layer. The data input layer starts with Hadamard gates on every qubit $q_i$ (where $i\in\left\{1,...,n\right\}$), then proceeds with $Ry$ gates rotating for $\arctan\left(x_i\right)$ and finishes analogously with $R_z$ rotating for $\arctan\left(x^2_i\right)$, where $x_i$ denotes the scalar data inputs. The variational layer starts with cyclic CNOTs where every qubit $q_i$ iteratively acts as the control for the $q_{i+k \mod n}$-th qubit, where $k\in\left\{1,2\right\}$.} for the low (moderate) complexity datasets~\cite{9747369}. The shape of $h$ and $c$ inside the (Q)LSTM cell are determined by concatenation $v_t$ of the previous hidden state $h_{t-1}$ and current input vector $x_t$.

\subsection{DisCoCat Implementation}
For implementing the DisCoCat approach, we follow the procedure stated in Sec.~\ref{subsec:DisCoCat}. In doing so, the first key component is a suitable pregroup parser. As no stable pregroup parsers exists at the time of conducting this research, we use the common, closest suited substitution: A CCG parser~\cite{yeung2021ccg}. In our implementation we used the BobCatParser from the Lambeq~\cite{kartsaklis2021lambeq} python library for this task. For translating the DisCoCat diagrams into quantum circuits, we follow~\cite{10.1613/jair.1.14329} by choosing these ansätze: Single qubit words are embedded with the standard Euler parameterization (i.e., $R_x(\theta_3)R_z(\theta_2)R_x(\theta_1)\ket{0}$) and words spanning over $d>1$ qubits are embedded with $d$ many IQP layers~\cite{havlivcek2019supervised}, which initially apply a Hadamard gate to each qubit and then apply parameterized, controlled $Rz$ gates in a chain-like pattern iteratively for all neighboring wires.

As the only available software implementation for DisCoCat\footnote{For details, see \href{https://github.com/ichrist97/qnlp\_finance}{https://github.com/ichrist97/qnlp\_finance}.} merely supported CPU-based circuit simulation (opposed to the available GPU-support for the (Q)LSTM implementation) lead to major wallclock time issues, we restricted the DisCoCat approach to binary classification by dropping the neutral data class data. While this generally makes the results less comparable, its impact is arguably negligible in the encountered case where QLSTM outperforms DisCoCat. The fundamental benefit of this is, that $\ket{s}$ can be represented with a single qubit, as it only needs to carry binary information hence requiring the least possible computational effort.

\section{Evaluation}
\label{sec:evaluation}
To evaluate the applicability of QLSTMs and DisCoCat to sentiment analysis in finance, we employ both approaches as specified in Sec.~\ref{sec:methodology} on the generated datasets using an 80/10/10 train/validation/test split and employing the binary cross entropy loss. As displayed in Fig.~\ref{fig:eval:qlstm-training}, the classical and the QLSTM are capable of learning the data in merely 20 epochs with validation accuracies over 80\%, hence showing very promising performance. For the moderate complexity dataset, much longer training times appear to be necessary, especially for the QLSTM, where it takes about 100 epochs before training progress starts to manifest, indicating possible barren plateaus. A more extensive hyperparameter search might circumvent this issue, but as the simulation already took 10 hours with a small sized cloud compute approach, classical circuit simulation appears to become a substantial bottleneck. In course of the conducted hyperparameter optimization, this effect appears to be strongly correlated with the number of ansatz layers employed, which is sensible, as an increased number of layers leads to a bigger number of parameters -- which is a natural cause of barren plateaus~\cite{skolik2021layerwise}. This correspondence indicates a time/quality trade-off which should be investigated more closely in future work.

For the DisCoCat approach, only the low complexity data was used, as the hardware requirements were substantially higher with 100 epochs taking roughly 82 hours of wallclock time with the same setup used for the QLSTM evaluation. Examining the results of DisCoCat displayed in Fig.~\ref{fig:eval:discocat-training}, we can observe that while a slight trend of improvement in the loss and accuracy is visible over the course of all 100 epochs, it is barely better than random guessing, which would achieve a 50\% accuracy. This indicates that QLSTMs are a lot faster to train than DisCoCat based approaches for realistic data sizes and complexities when using the employed CPU based circuit simulator. However, we argue against generalizing these results towards ''QLSTMs outperform DisCoCat``, as a clear learning progress of DisCoCat is observable. Extrapolating from similar studies~\cite{10.1613/jair.1.14329}, we expect reasonably accurate results for DisCoCat given a more efficient software implementation.

\begin{figure}
    \centering
    \includegraphics[width=\columnwidth]{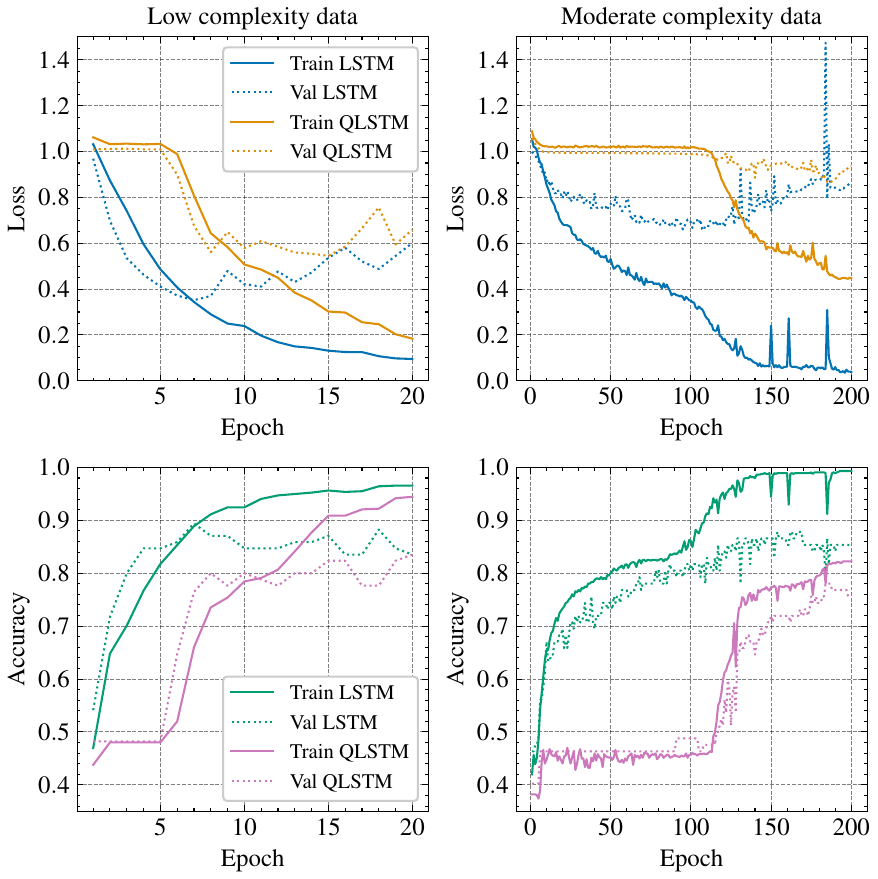}
    \caption{Training curve for LSTM and QLSTM on both datasets.}
    \label{fig:eval:qlstm-training}
\end{figure}

\begin{figure}
    \centering
    \includegraphics[width=\columnwidth]{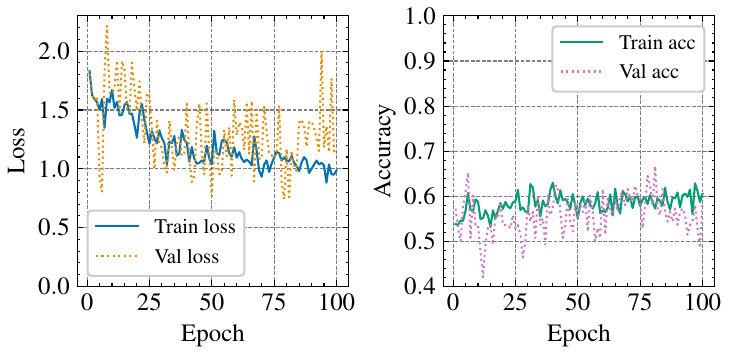}
    \caption{Training curve for DisCoCat on low complexity data.}
    \label{fig:eval:discocat-training}
\end{figure}

\section{Conclusion}
\label{sec:conclusion}
In sight of current QC hard- and software limitations, QLSTMs seem to be better suited when solving QNLP tasks like sentiment analysis on the employed datasets, which appear to be a lot more realistic than those used in existing proofs of concept for both approaches. Notably, the employed ChatGPT-based data generation approach enabled producing realistic data, while preserving necessary grammatical correctness, which facilitates more practical testing. A future possibility to overcome the experienced circuit simulation bottleneck might be the support of GPU based circuit simulation for the employed implementation using pennylane and Lambeq.

\section*{Acknowledgment}
This paper was partially funded by the German Federal Ministry for Economic Affairs and Climate Action through the funding program "Quantum Computing -- Applications for the industry" based on the allowance "Development of digital technologies" (contract number: 01MQ22008A). Furthermore, this paper was partially funded by the German Federal Ministry of Education and Research through the funding program "Network of interdisciplinary education and training concepts in quantum technologies" based on the allowance "Quantum technologies -- from fundamentals to market" (contract number: 13N16000).

\bibliographystyle{unsrt}  
\bibliography{main} 

\end{document}